\label{key}

\documentclass[preprint,12pt]{elsarticle}

\usepackage{amssymb}
\usepackage{stfloats}
\usepackage{caption}

\usepackage{amsopn}
\usepackage{amsmath}
\usepackage{float}
\usepackage{makecell,multirow,diagbox}
\usepackage{url}

\usepackage{subfigure}
\if CLASSOPTIONcompsoc
\usepackage[caption=false,font=normalsize,labelfont=sf,textfont=sf]{subfig}
\else
\usepackage[caption=false,font=footnotesize]{subfig}
\fi
\usepackage{caption}
\usepackage{subfigure}

\biboptions{numbers,sort&compress}

\journal{}

\begin{document}

\captionsetup[figure]{labelfont={bf},labelformat={default},labelsep=period,name={Fig.}}

\begin{frontmatter}



\title{A Deep Retinal Image Quality Assessment Network with Salient Structure Priors}


\author{Ziwen Xu}
\author{Beiji Zou}
\author{Qing Liu}
\address{School of Computer Science and Engineering, Central South University}

\begin{abstract}
Retinal image quality assessment is an essential prerequisite for diagnosis of retinal diseases. Its goal is to identify retinal images in which anatomic structures and lesions attracting ophthalmologists’ attention most are exhibited clearly and definitely while reject poor quality fundus images. Motivated by this, we mimic the way that ophthalmologists assess the quality of retinal images and propose a method termed SalStructuIQA. First, two salient structures for automated retinal quality assessment. One is the large-size salient structures including optic disc region and exudates in large-size. The other is the tiny-size salient structures which mainly include vessels. Then we incorporate the proposed two salient structure priors with deep convolutional neural network (CNN) to shift the focus of CNN to salient structures. Accordingly, we develop two CNN architectures: Dual-branch SalStructIQA and Single-branch SalStructIQA. Dual-branch SalStructIQA contains two CNN branches and one is guided by large-size salient structures while the other is guided by tiny-size salient structures. Single-branch SalStructIQA contains one CNN branch, which is guided by the concatenation of salient structures in both large-size and tiny-size. Experimental results on Eye-Quality dataset show that our proposed Dual-branch SalStructIQA outperforms the state-of-the-art methods for retinal image quality assessment and Single-branch SalStructIQA is much light-weight comparing with state-of-the-art deep retinal image quality assessment methods and still achieves competitive performances.
\end{abstract}


\end{frontmatter}


\section{Introduction}
Retinal images are the most common modality used in the diagnosis of retinal diseases such as diabetic retinopathy, age-related macular degeneration and glaucoma etc. High quality retinal images make the salient structures, such as optic disc, vessels, and lesions etc clearly visible, which help ophthalmologists make correct clinical decisions efficiently. On the contrary, low quality images may disturb the judgments. For example, in over-exposed retinal images, the brightness and the range of colours are limited and the contrasts between target regions (such as vessels and lesions etc.) and the background are low, which may confuse the ophthalmologists when accessing the fundus. Furthermore, for computer-aided retinal image analysis systems, low quality retinal images would be a catastrophe. According to \cite{macgillivray2015suitability}, approximately $25\%$ of the retinal images are not suitable for diagnosis due to their poor quality. Thus retinal image quality assessment (Retinal-IQA) is desired. In clinical, it is performed by a trained optometrist manually, which heavily depends on the operator's experience and is time-consuming. To improve the efficiency of retinal image acquisition, automated Retinal-IQA becomes necessary.

The goal of image quality assessment (IQA) is to grade images in terms of quality measures. Generally, IQA methods can be categorized into three groups based on the presence of reference images: full-reference (FR), reduced-reference (RR) and no-reference (NR) \cite{wang2006modern}. Within the context of Retina-IQA, retinal reference images are absent, thus the task of Retinal-IQA belongs to NR-IQA. It is always formulated as either a two-class, i.e., \lq Good\rq~and \lq Reject\rq~\cite{zago2018retinal} or three-class \cite{fu2019evaluation} (i.e. \lq Good\rq~, \lq Usable/Acceptable\rq~and \lq Reject\rq) classification task. As two-class classification is too rigorous and retinal images with a few artifacts can still be usable for ophthalmologists, the prevalence of Retina-IQA is dominated by three-class Retina-IQA methods. 

Early Retinal-IQA methods grade the image quality via solving the problems of how to measure the similarity to good quality retinal images or how to characterise quality distortions. In \cite{lee1999automatic, lalondey2001automatic}, template matching is adopted for Retinal-IQA, in which
intensity histogram and intensity-gradient histogram are calculated from a set of good quality retinal images as templates. In \cite{dias2014retinal, wang2015human}, quality distortions are characterised by colour and illumination distortion, focus fluctuation and contrast sensitivity, with which a classifier is trained for retinal image quality grading.  Although those methods are simple and intuitive, both templates based on histograms and features manually designed for quality distortions are insufficient to encode quality measures with high discriminative ability, which limits the performances of Retinal-IQA.





Driven by the successes achieved by deep convolutional neural networks (CNNs) in the field of computer vision, Retinal-IQA methods have updated to modern CNNs-based. In \cite{zago2018retinal}, Inception v3 \cite{szegedy2016rethinking} pre-trained on natural scene image dataset ImageNet \cite{deng2009imagenet} is fine-tuned in an end-to-end way for Retinal-IQA, which takes colour retinal images as input. To make use of both deep features and hand-crafted features, \cite{yu2017image} proposed a two-stage learning method. In detail, deep features are learned in the first stage via fine-tuning an AlexNet \cite{krizhevsky2012imagenet}. With the deep features and hand-crafted features based on a saliency map, a SVM is trained for quality classification in the second stage. To further exploit powerful representation for Retinal-IQA, complex frameworks are proposed. Fu et al. propose a framework, called Multiple Colour-space Fusion Network (MCF-Net) \cite{fu2019evaluation}, in which three parallel CNN branches are unified in one framework to learn complementary informative contexts from three different colour-spaces, i.e. RGB, HSV and Lab colour space, for Retinal-IQA. In \cite{shen2018multi}, Shen et al. claim that auxiliary tasks contribute to Retina-IQA and propose a multi-task framework, named MFIQA. MFIQA \cite{shen2018multi} consists of a ResNet50 \cite{he2016deep} for optic disc and macular detection, a VGG16 \cite{simonyan2014very} for refinement of optic disc and macular location and two encoders to encode the global retinal images in Cartesian coordinate system and the local optic disc windows in polar coordinate system for images quality classification and three relevant classification tasks, i.e., artifact, clarity and field definition. It is further improved by introducing domain invariance and interpretability in \cite{shen2020domain}. Although learning multiple CNN branches jointly like \cite{fu2019evaluation} and multiple CNNs independently like \cite{shen2018multi, shen2020domain} exploit rich features and facilitate classification, the parameters to be optimised rapidly multiply. Furthermore, in \cite{shen2018multi, shen2020domain}, learning auxiliary tasks requires extra annotated data, which are always expensive.

\begin{figure*}[!ht]
	\centering
	\includegraphics[width=0.8\textwidth]{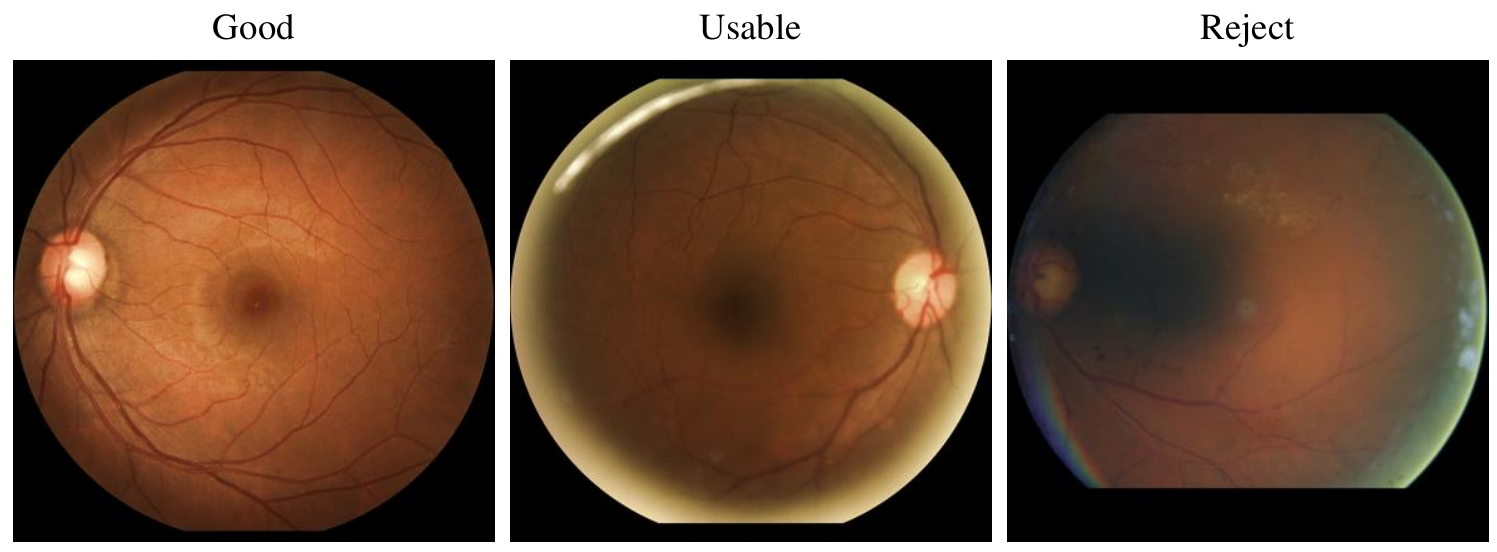}
	\caption{Examples for retinal images with different quality grades. From left to right, the quality grade is \lq Good\rq, \lq Usable\rq~and \lq Reject\rq, respectively.  }
	\label{fig:msss_vessel_compare}
\end{figure*}

Going back to the basics, the goal of capturing retinal images by fundus camera is to assist ophthalmologists to assess anatomic structures on fundus in a non-invasion manner. From the view of saliency detection, those anatomic structures such as optic disc in large size and vessels in tiny size attract most of attention of ophthalmologists. Thus a good quality retinal image is required to exhibit those salient structures clearly and definitely. To intuitively illustrate this, we show three examples from Eye-Quality dataset \cite{fu2019evaluation} in Fig. \ref{fig:msss_vessel_compare}. They are graded as \lq Good\rq, \lq Usable\rq~and \lq Reject\rq~respectively. Obviously, for good quality retinal image as shown in the left column in Fig. \ref{fig:msss_vessel_compare}, the optic disc region and vessels are clearly visible. For usable quality retinal image as shown in middle column in Fig. \ref{fig:msss_vessel_compare}, the optic disc region and most of the vessels are clearly visible except for vessel extending to the boundary of field of view which are distorted by slight uneven illumination. For retinal image labelled as \lq Reject\rq, due to the serious uneven illumination, the optic disc region and vessels are difficult to be distinguished.

\begin{figure*}[!ht]
	\centering
	\includegraphics[width=0.9\linewidth]{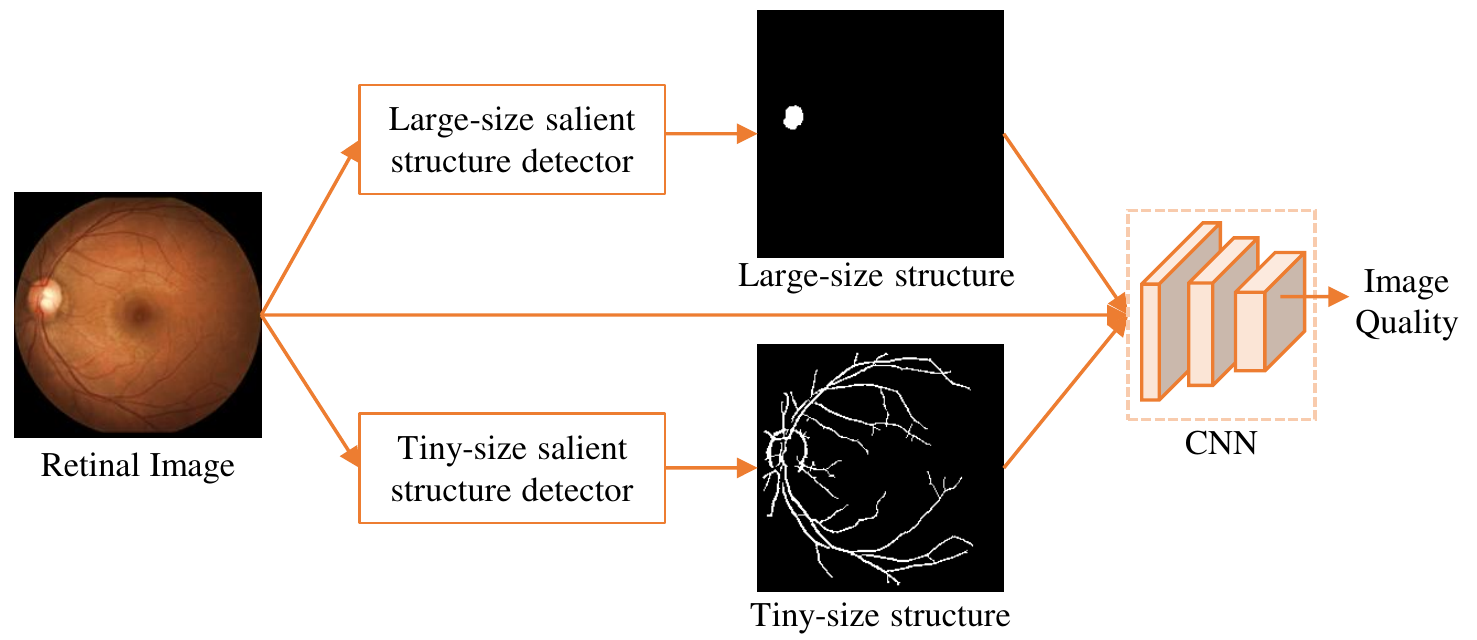}
	\caption{The flowchart of our proposed SalStructIQA.}
	\label{fig:flow}
\end{figure*}

Based on above observation, we argue that the clarity of salient structures in both large-size and tiny-size significantly contribute to the image quality assessment. Usually, more clarity that salient structures in retinal images are implies higher quality that retinal images have. To this end, we propose to incorporate the salient structure priors with deep CNN, named SalStructIQA. Fig. \ref{fig:flow} illustrates the flowchart of our proposed SalStructureIQA. It involves three components: (1) large-size salient structure detector which aims to detect salient structures in large-size such as optic disc region, large-size hard exudates and soft exudates etc., (2) tiny-size salient structure detector which aims to detect vessels, (3) a deep classification network which takes the retinal image together with the salient structures as input for deep feature learning and quality classification. Accordingly, we develop two deep CNN architectures to incorporate salient structure priors with the CNNs. One contains two parallel CNN branches for deep feature learning, which learns deep features from retinal image with large-size salient structures and retinal image with tiny-size salient structures separately, then fuses the features for quality assessmentenforce. We term it as Dual-branch SalStructIQA. The other only contains a single CNN branch which first fuses the retinal image and salient structures via concatenation, then learns deep features for quality classification. We term it as Single-branch SalStructIQA. We demonstrate the effectiveness of our proposed priors and two CNN architectures on public dataset Eye-Quality \cite{fu2019evaluation}. And experimental results show that our method outperforms the state-of-the-arts.

In summary, the contributions of this paper can be summarized as follows:
\begin{itemize}
	\item We propose two salient structure priors in both large-size and tiny-size for retinal image quality assessment. They enable the CNN to pay more attention to informative regions with high contrast to background and learn discriminative features, which further facilitate to retinal image quality assessment.
	\item We propose two ways to incorporate our proposed salient structures with deep CNN and develop two deep CNN architectures termed Dual-branch SalStructIQA and Single-branch SalStructIQA for retinal image quality assessment. Experimental results on Eye-Quality \cite{fu2019evaluation} show that the Single-branch one is able to achieve competitive performances comparing with the state-of-the-arts but the scale of parameters to be optimised is much less. The Dual-branch one outperforms the state-of-the-arts as well as the Single-branch.
\end{itemize}

The remainder of this paper is organized as follows. Section \ref{sec:relatedwork} reviews related works on Natural-IQA and Retinal-IQA. Section \ref{sec:proposedmethod} details our proposed quality assessment method. Section \ref{sec:exprst} gives the experimental results and analysis and Section \ref{sec:conclusion} concludes this paper.

\section{Related work}
\label{sec:relatedwork}
Following the evolution of natural scene image quality assessment (Natural-IQA), retinal image quality assessment methods are developed from the two-step framework with hand-crafted features to the end-to-end framework based on CNN in general. As retinal image quality assessment is closely relative to natural scene image quality assessment, in this section, we first review previous works on Natural-IQA briefly, then elaborate preceding Retinal-IQA methods.

\subsection{Natural-IQA}
Based on the type of extracted features, Natural-IQA methods can be briefly classified into two groups:  hand-crafted feature based methods and CNN-based methods. The hand-crafted based approaches mainly focus on designing effective hand-crafted features while the CNN-based methods automatically learn image features by designing deep network. 

\textbf{Hand-crafted feature based methods.} These methods usually assume that pristine natural images possess potential statistic regularities and these regulations will be changed with the emergence of distortion \cite{moorthy2011blind}. For example, \cite{moorthy2011blind} proposes a two-stage framework called DIIVINE for quality assessment, which consists of distortion classification and distortion-specific quality estimation. BLINDS-II \cite{Saad2012Blind} utilizes DCT coefficients to model statistics of images, and then combines the estimated parameters of the model with Bayesian approach for quality score prediction. BRISQUE \cite{mittal2012no} adopts local normalization brightness coefficients for image quality. To better improve the performance, a new NBIQA \cite{Fu2019A} simultaneously considers the features from both the spatial and transform domains. Although NBIQA \cite{Fu2019A} performs better, it is also relatively time-consuming due to the complexity of extracted features. In addition, \cite{li2019blind} designs a joint NSS model based on joint log-contrast statistics for Natural-IQA, which utilizes two natural image properties, i.e., the non-Gaussian statistics and image correlations across spatial and direction. As a whole, the hand-crafted feature based methods heavily rely on the artificial features design, which is too complex to achieve efficient and appropriate features.

\textbf{CNN-based methods.} Due to the remarkable achievements of CNN in image classification tasks, researchers have introduced the deep learning method into Natural-IQA. In \cite{2014Convolutional},  the CNN only containing one convolutional layer is proposed for Natural-IQA. To better characterize features, \cite{ma2017end} proposes a multi-task end-to-end optimized network for quality prediction called MEON. Similar to DIIVINE \cite{moorthy2011blind}, MEON focuses on two sub-task, distortion identification and quality prediction. In \cite{ma2017dipiq,liu2017rankiqa}, rank learning is adopted for Natural-IQA. In rankIQA \cite{liu2017rankiqa}, the traditional image processing method is firstly utilized to generate ranked images which are then used to train a siamese network for ranking. Subsequently, the branch of trained network is fine-tuned on IQA dataset for predicting the absolute quality score. In dipIQ \cite{ma2017dipiq}, a large number of quality-discriminable image pairs are used to train a RankNet \cite{Christopher2005Learning}, generating an inferred image quality. However, the above CNN-based methods ignore the salient structures attracting human attention, which have been demonstrated to be important for image quality assessment \cite{yan2018two,yang2019sgdnet,yang2020deep}. \cite{yan2018two} proposes a two-stream network for Natural-IQA, which includes an image stream focusing on gray information and a structure stream focusing on image gradient details. Considering that image saliency can present image structure to some extent, \cite{yang2019sgdnet} proposes a saliency-guided Natural-IQA network called SGDNet, which is trained with saliency ground truth and quality label. Similar to approach \cite{jia2018saliency}, SGDNet regards the saliency map as local weighted function. Specifically, the predicted saliency map is fused with original features in the form of spatial attention mask, perceptually consistent to quality prediction. \cite{yang2020deep} proposes a deep feature importance awareness method to evaluate image quality named DFIA. They integrate the Squeeze-and-Excitation (SE) block \cite{hu2018squeeze} into deep IQA network, which emphasizes the deep features of image distortion and salient objects related to IQA.   

\subsection{Retinal-IQA}
Retinal-IQA is an important task for the diagnosis of retinal diseases. However, Natural-IQA methods may not perform well in Retinal-IQA. More specifically, the hand-crafted based methods for Natural-IQA mainly rely on a hypothesis that pristine natural images present certain statistic regularities which will be disturbed by distortions. However, the statistics of natural images are not consistent with those of retinal images. In addition, several CNN-based methods for Natural-IQA usually assign the global quality score for all image patches cropped from the same image, resulting the label noise problem. This strategy is also not suitable for Retinal-IQA. Therefore, it is essential to design effective methods for Retinal-IQA. Similar to Natural-IQA, the current researches in Retinal-IQA can be roughly grouped to two categories: hand-crafted based methods and CNN-based methods.   

\textbf{Hand-crafted based methods.} Early methods such as \cite{lee1999automatic,lalondey2001automatic} for Retinal-IQA pursue a template for good quality images and grade the image quality via template matching while \cite{davis2009vision,dias2014retinal} extract statistical characteristics such as focus, lumination and contrast features etc. and train a classifier for image quality assessment. Based on the assumption that the more vessels the better retinal image quality, \cite{usher2003automated} first calculates the area of vessels to define image clarity, in which matched filtering and region segmentation algorithm are applied to segment the vessel structures. In \cite{kohler2013automatic}, Kohler et al. assume that patches located on vessel edges are more sensitive to image blur and noise, and propose to use the variance of vessels in patches to discriminate good retinal images from poor quality retinal images. In \cite{hunter2011automated}, retinal image quality is graded by comparing the vessel clarity and contrast between the macular area and background. From the perspective of human visual system (HVS) characteristics, \cite{wang2015human} proposes to apply HVS features, including multi-channel sensation, just noticeable blur, and the contrast sensitivity measure, to estimate quality. However, the limitation of these hand-crafted features may still lead to performance bottlenecks. 

\textbf{CNN-based methods.} Motivated by the success of CNN in computer vision tasks, CNN has been introduced to Retinal-IQA. Several methods for Retinal-IQA are either training a shallow classification CNN network from scratch \cite{tennakoon2016image} or fine-tuning existing classification networks, such as AlexNet \cite{krizhevsky2012imagenet}, ResNet \cite{he2016deep} and Xception \cite{chollet2017xception}, from pre-trained models \cite{zago2018retinal,saha2018automated,raj2020multivariate}. To make use of the complementary of different colour spaces, Fu et al. \cite{fu2019evaluation} propose a multiple colour-space fusion network, in which multiple parallel CNN branches are used to learn features from different colour-spaces. Although \cite{fu2019evaluation} is able to capture richer information such that the classification performances are improved significantly, the scale of parameters rapidly multiply. Instead, Yu et al. \cite{yu2017image} propose to combine salient maps produced by \cite{achanta2009frequency} and deep features extracted by Alexnet \cite{krizhevsky2012imagenet} and train an SVM for retinal image quality assessment. In \cite{shen2018multi}, Shen et al. claim that auxiliary tasks contribute to Retinal-IQA and propose a multi-task framework named MFIQA, in which a ResNet50 \cite{he2016deep} is trained for the detection of optic disc and macular and a VGG16 \cite{simonyan2014very} is trained for refinement of optic disc and macular location and two encoders are used to encode the entire retinal images in Cartesian coordinate system and the local optic disc windows in polar coordinates systems for images quality classification. Additionally, three relevant classification tasks, i.e., artifact, clarity and field definition are proposed to assist the learning of quality assessment. To further improve the performances, domain invariance and interpretability are introduced in \cite{shen2020domain}. Learning auxiliary tasks do facilitate Retinal-IQA, but requires extra annotated data, which are expensive.

\section{Methodology}
\label{sec:proposedmethod}
The purpose of photographing for retina fundus is to provide an invasive way for ophthalmologist to assess the fundus. Commonly ophthalmologists pay more attention to informative structures. Among them, two kinds of informative structures attract most attentions from ophthalmologists: structures in large-size such as OD and large-size hard exudates and soft exudates etc. and structures in tiny-size, i.e., vessels. Whether these salient structures are clearly and definitely exhibited plays an important role in determining whether retinal images have good quality for ophthalmologists. To mimic the quality assessment procedure of ophthalmologists, we propose a deep retinal image quality assessment method termed SalStructIQA, which first detects two kinds of salient structures, then incorporates them with deep CNN for quality grading. Fig \ref{fig:flow} illustrates the flowchart of our proposed SalStructIQA. It consists of three components: large-size salient structure detector, tiny-size salient structure detector and deep CNN classifier. In the following, we will detail each of them.


\subsection{Large-size salient structure detector}
Within the context of retinal images, large-size structures mainly consists of OD and large-size exudates etc. They are in bright yellow. In good quality retinal images, those structures are easy to be recognised. On the contrast, in poor quality retinal images, those structures are difficult to be recognised. The major reason is that the contrasts of those large-size salient structures in different quality images are different. Particularly, for good quality retinal images, large-size structures naturally have high contrast to their surrounding regions as images are usually captured with even illumination. For poor retinal images, they are usually captured with uneven illumination. Defect regions with strong illumination exhibit high contrast while the contrasts of those bright yellow structures with weak illumination are suppressed. To this end, we propose to detect large-size salient structures via contrast for retinal image quality assessment. It is estimated by Difference of Gaussian (DoG) operator, which can be expressed as:
\begin{equation}
\begin{array}{ll} 
DoG(x, y,\sigma, \rho )  & =\frac{1}{2\pi}\left[\frac{1}{\sigma^2} \mathrm{e}^{-\frac{x^2+y^2}{\sigma^2}} - \frac{1}{(\sigma/\rho)^2} \mathrm{e}^{-\frac{x^2+y^2}{(\sigma/\rho)^2}}\right]\\
& =G(x,y,\sigma) - G(x,y, \sigma/\rho)
\end{array} \;,
\end{equation}
where $\sigma$ is the standard deviation of Gaussian and $\rho$ is a constant which is larger than 1. From the view of frequency domain, the DoG operator is a band-pass filter and $\rho$~is the bandpass width. Considering that the salient structures are diverse in scale, we propose to combine multiple DoGs with $K$ different bandpass widths for contrast estimation:
\begin{equation}
\label{eq:MSDoG}
\begin{array}{ll} 
MultiDoG(x,y) & = \sum_{k=1}^{K} \left\{DoG(x, y,\sigma_{k},\rho^{k-1} ) - DoG(x, y,\sigma_{k},\rho^{k} ) \right\}\\
& =G(x,y,\sigma) - G(x,y, \sigma/\rho^{K})
\end{array} \;,
\end{equation}
\begin{figure}[!ht]
	\centering
	\includegraphics[width=0.95\linewidth]{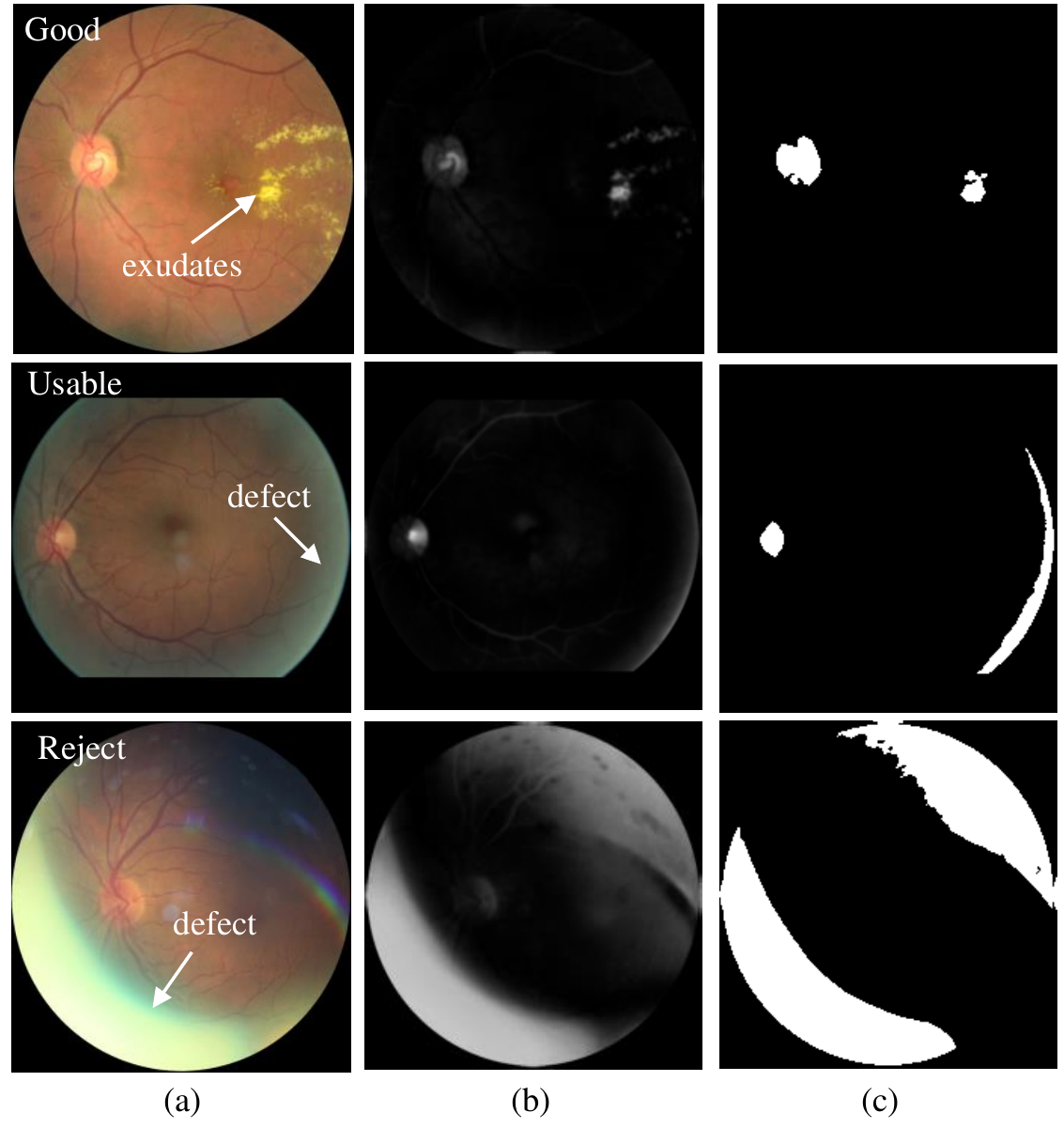}
	\caption{Examples for the probability map of large-size salient structures and the masks of detected large-size salient structures. }
	\label{fig:lsmap}
\end{figure}
Here we directly set $\sigma$ be a large constant to maintain most of the frequency and filter out the high frequency noises. Thus the first item in Eq. (\ref{eq:MSDoG}) is approximated to a low-pass filter with a large cut-off frequency. For computational simplicity, we follow \cite{achanta2009frequency} and use a $5\times 5$ Gaussian filter to approximate it in spatial domain. For the second item in Eq. (\ref{eq:MSDoG}), we set $K$ be infinity, thus it is approximated to a low-pass filter at zero frequency. More specifically, with the colour retinal image $\mathbf{I}\in \mathcal{R}^{H\times W \times 3}$ in RGB colour space, we first transform it into \textit{Lab} colour space as \textit{Lab} is designed to approximate human vision. We denote the \textit{Lab} image as $\mathbf{I}^{Lab}$. Then in spatial domain, we estimate the contrast via: 
\begin{equation}
\mathbf{R}_{contrast}(x,y) = ||\mathbf{I_{\mu}}^{Lab} - \mathbf{I}_{lp}^{Lab} (x,y)||_2\;,
\end{equation}
where $\mathbf{I_{\mu}}$ is the global mean vector of $\mathbf{I}^{Lab}$,  $\mathbf{I}_{lp}^{Lab} (x,y)$ is the corresponding filter response at location $(x, y)$ with the $5\times 5$ Gaussian filter, and $||\cdot||_2$ is the Euclidean distance. Finally, we rescale the contrast map into $[0, 1]$ via max-normalisation, and obtain the probability map of large-size salient structures denoted by $\mathbf{P}_{LS}$:
\begin{equation}
\mathbf{P}_{LS}(x,y) = \frac{\mathbf{R}_{contrast}(x,y)}{\max_{(x,y)\in \mathcal{R}^{H\times W}} \{\mathbf{R}_{contrast}(x,y)\}}\;.
\end{equation}
By thresholding $\mathbf{P}_{LS}$, we obtain the mask of detected large-size salient structures, and we denote it as $\mathbf{M}_{LS}$.

Fig. \ref{fig:lsmap} illustrates three retinal images from Eye-Quality dataset \cite{fu2019evaluation}, their corresponding probability map $\mathbf{P}_{LS}$ of large-size salient structures and binarized results. From top to bottom, retinal images are graded as \lq Good\rq, \lq Usable\rq~and \lq Reject\rq~by experts respectively. Obviously, their detection results of large-size salient structures are significantly different: (1) in retinal image graded as \lq Good\rq, large-size salient structures are well highlighted in $\mathbf{P}_{LS}$; (2) in retinal image graded as \lq Reject\rq, defects are well highlighted in $\mathbf{P}_{LS}$ while the salient structures are suppressed; (3) in retinal image graded as \lq Usable\rq, although slight detect is highlighted in $\mathbf{P}_{LS}$, its adverse effect is weak and the salient structure is still well highlighted. These imply that the detection results of large-size salient structures do contribute to image quality grading. 

\subsection{Tiny-size salient structure detector}
Vessels are important structures in retinal image. Their abnormal morphology is a significant manifestation for eye diseases such as diabetic retinopathy and age related macular degeneration. Although they are in tiny size, they are salient for ophthalmologists. Thus we propose to detect those tiny-size salient structures for retina image quality assessment.  
\begin{figure}[!ht]
	\centering
	\includegraphics[width=0.95\linewidth]{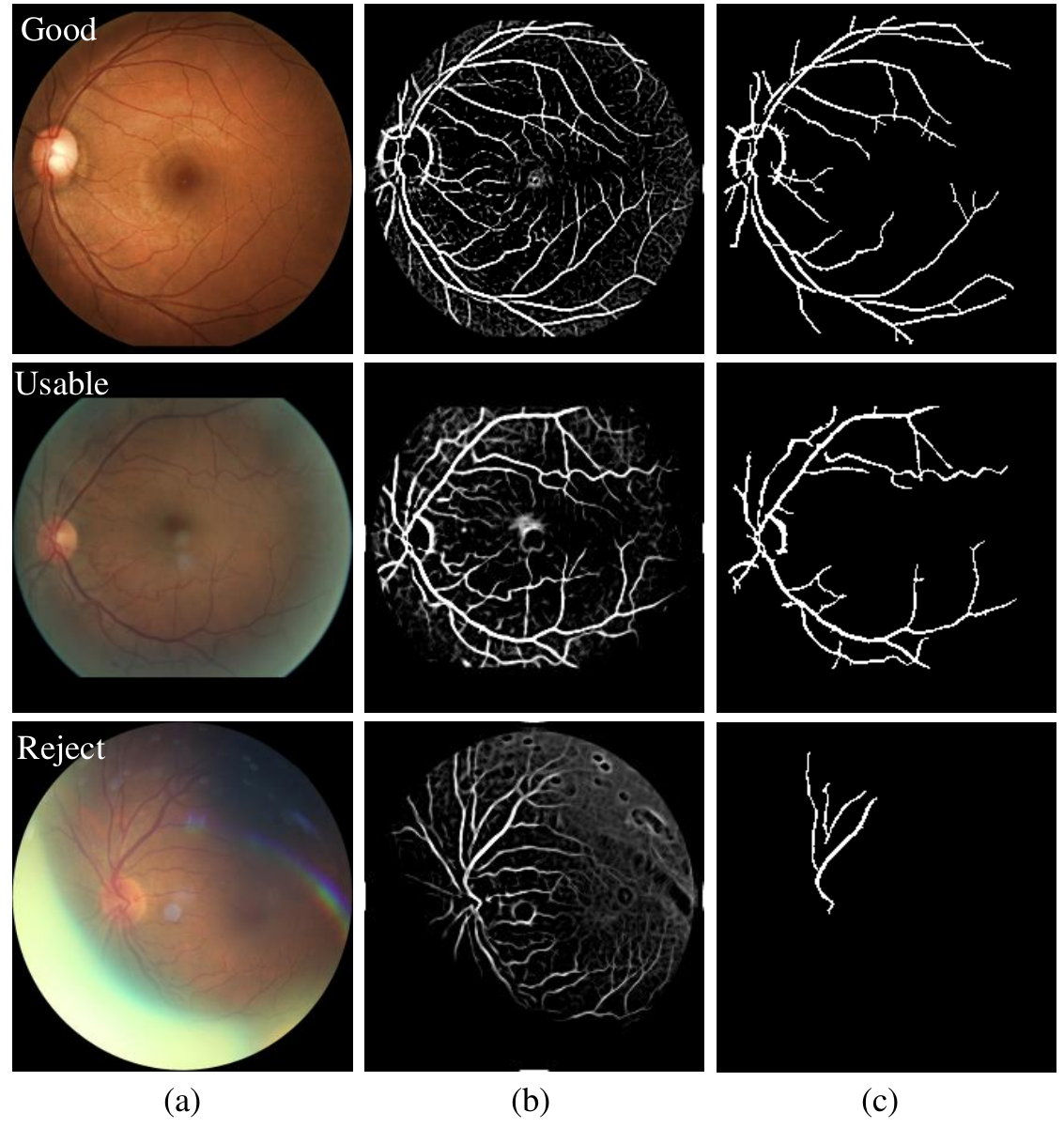}
	\caption{Examples for the standardized line response maps and masks of detected tiny-size salient structures.}
	\label{fig:vesselmap}
\end{figure}

Here we directly adopt multi-scale line detector \cite{nguyen2013effective} as our tiny-size salient structure detector due to its simplicity. Multi-scale line detector \cite{nguyen2013effective} treats the vessels as line-shaped and assumes that there are high intensity contrasts between pixels along vessel line and rest pixels among multi-scale windows. As the vessels is darker than the background in the gray channel $\mathbf{I}^{Gray}$ of a retinal image, the multi-scale line detector is performed on its inverted gray channel to highlight the vessels. For a pixel at location $(x, y)$, the multi-scale line detector measures the line response by:
\begin{equation}
	\mathbf{R}_{line}(x, y) = \frac{1}{|S|}\left(\sum_{s\in {S}} \max_{\theta\in\Theta} \left(Avg(L_{x, y}^{s,\theta})-Avg(\Omega^{s}_{x,y})\right) \right)\;,
\end{equation}
where $S$ is the set of window size, and $|S|$ is the cardinality of set $S$, $\Theta$ is the set of line direction, and $\Omega^{s}_{x,y}$ is the local window of size $s\times s$ centred on $(x, y)$, and $L_{x, y}^{s,\theta}$ is the pixel set of line along direction $\theta$ passing location $(x, y)$ in $\Omega^{s}_{x,y}$, and $Avg(\cdot)$ is an average function. We follow the setting of multi-scale line director \cite{nguyen2013effective} and set $S=\{1,3,5,7\}$ and $\Theta=\{0^{\circ}, 15^{\circ}, 30^{\circ}, \cdots, 165^{\circ}\}$. Since the contrast of vessels to other structures such as lesions is apparent in the original inverted gray image, multi-scale line detector \cite{nguyen2013effective} also utilizes the information of the inverted gray channel to enhance the line response:
\begin{equation}
	\mathbf{R}_{line}'(x, y) = \frac{1}{|S|+1}\left(\sum_{s\in {S}} \max_{\theta\in\Theta} \left(Avg(L_{x, y}^{s,\theta})-Avg(\Omega^{s}_{x,y})\right) + \left(1-\mathbf{I}^{Gray}(x,y)\right)\right)\;.
\end{equation}
Finally, we follow \cite{nguyen2013effective} and standardize the enhanced line response map $\mathbf{R}_{line}'$ via $Z-$score standardization. We denote the standardized line response map as $\mathbf{R}_{line}^{''}$. By thresholding the $\mathbf{R}_{line}^{''}$ with threshold value $0.56$, we obtain the mask of detected tiny-size salient structures and denote it as $\mathbf{M}_{TS}$.
\begin{figure}[!t]
	\centering
	\includegraphics[width=0.95\textwidth]{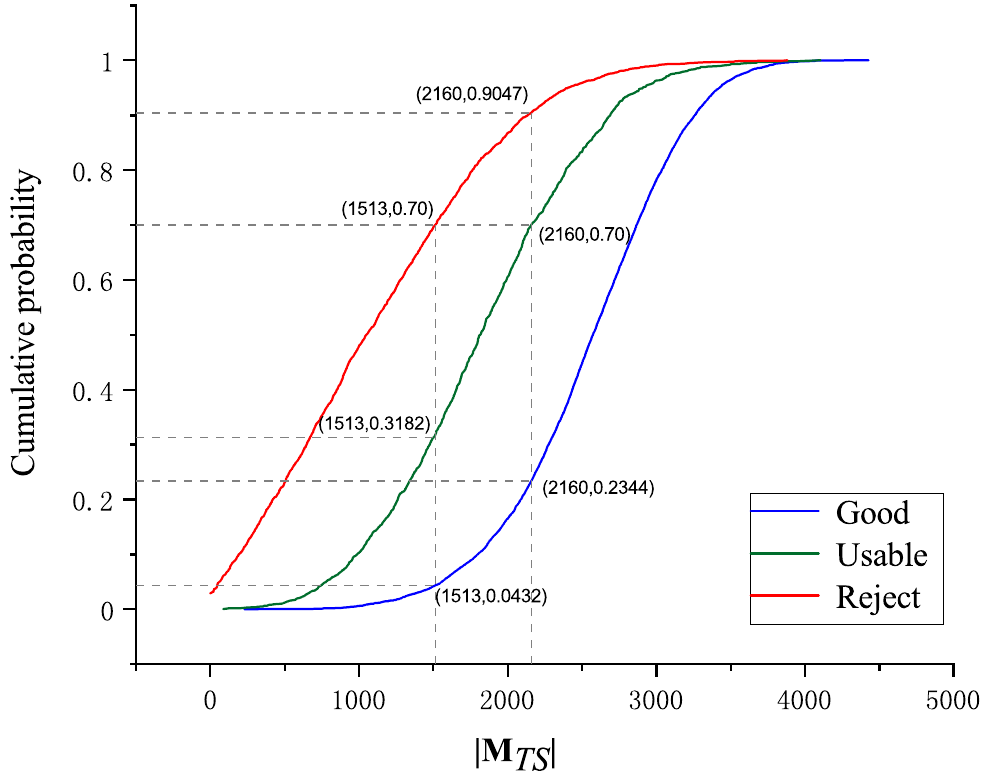}
	\caption{The cumulative distributions of tiny-size salient structures of each class.}
	\label{fig:vesselstatistic}
\end{figure}

Fig. \ref{fig:vesselmap} shows three examples of retinal images, the results of tiny-size salient structure detection. Obviously, the tiny-size salient structures in good quality retinal image are well detected by our proposed tiny-size salient structure detector. Most of the tiny-size salient structures in retinal image graded as \lq Usable\rq~quality are detected by our proposed detector. On the contrary, few tiny-size salient structures in retinal image graded as \lq Reject\rq~quality are detected by our proposed detector. These indicate that the density of detected tiny-size salient structures is an important cue for the quality classification of retinal images. To better illustrate this, we provide a statistic analysis on the training set of Eye-Quality \cite{fu2019evaluation}. For each retinal image, we account the number of pixels which are detected as tiny-size salient structures and denote it as $|\mathbf{M}_{TS}|$. Then for each quality level, we separately plot the cumulative distribution of $\mathbf{M}_{TS}$. As shown in Fig. \ref{fig:vesselstatistic}, there are 70\% images in class of \lq Reject\rq~whose $|\mathbf{M}_{TS}|$ is less than 1513 while the percentage in class of \lq Usable\rq~is 31.82\%. This implies that it is easy to discriminate the class of \lq Reject\rq~and \lq Usable\rq~according to $|\mathbf{M}_{TS}|$. There are 70\% images in class \lq Usable\rq~whose $|\mathbf{M}_{TS}|$ is less than 2160 while the percentage in class of \lq Good\rq~is 23.44\%. This implies that it is easy to discriminate the class of \lq Usable\rq~and \lq Good\rq~according to $|\mathbf{M}_{TS}|$.

  \begin{figure}[!ht]
 	\centering
 	\includegraphics[width=0.95\linewidth]{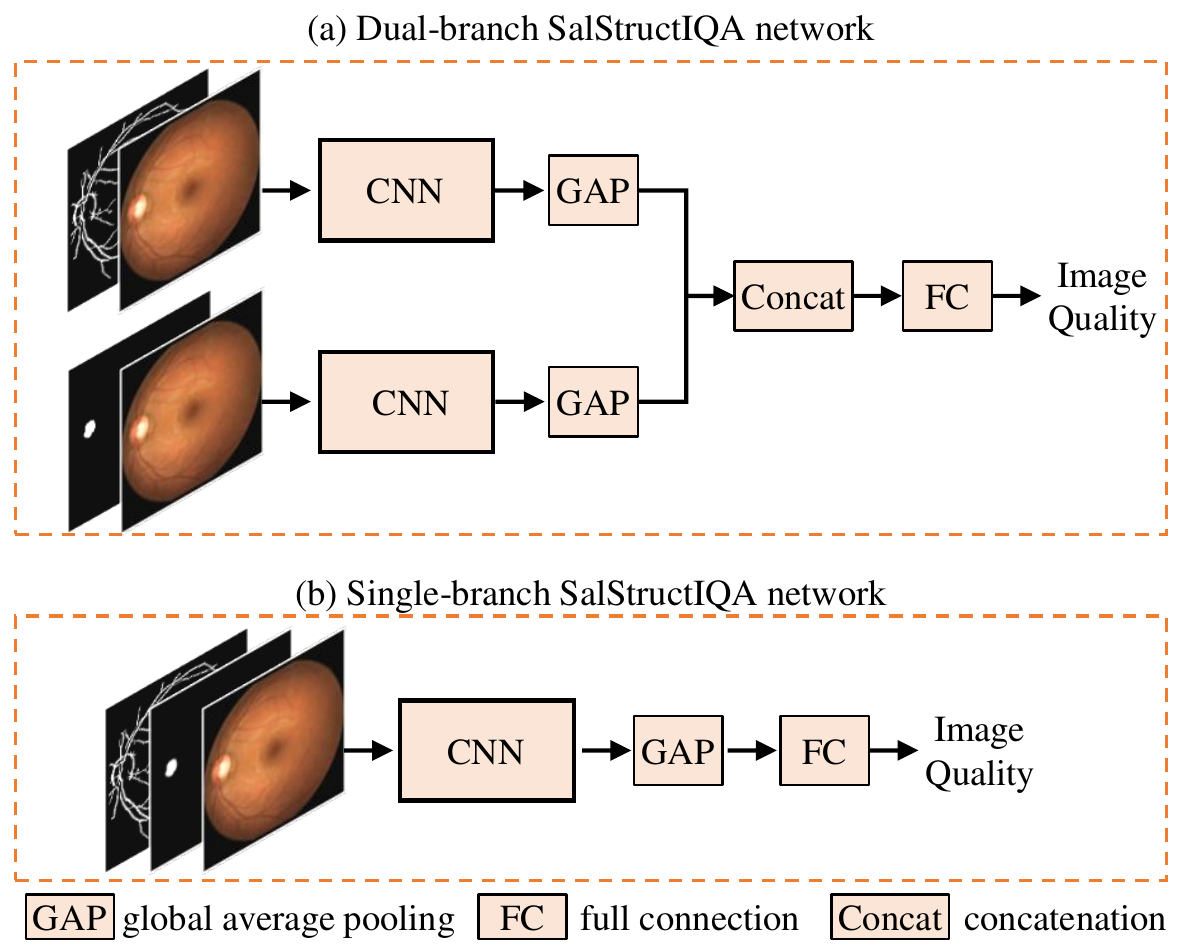}
 	\caption{The proposed deep CNN architectures incorporating with our proposed large-size salient structures and tiny-size salient structures.}
 	\label{fig:architecture}
 \end{figure}
\subsection{Network Architectures}
To incorporate two salient structure priors with convolutional neural network for retinal image quality grading, we implement a deep network, named SalStructIQA network. For these two priors, they can be separately concatenated with the RGB retinal image forming two tensors with four channels as shown in Fig. \ref{fig:architecture}(a) or jointly concatenated with the RGB retinal image forming one tensor with five channels as shown in Fig. \ref{fig:architecture}(b). This derives two versions of SalStructIQA: Dual-branch SalStructIQA and Single-branch SalStructIQA. In what follows, the RGB retinal image is denoted as $\mathbf{I}\in \mathcal{R}^{H\times W \times3}$, and the masks of large-size salient structures and tiny-size salient structures are denoted as $\mathbf{M}_{LS}\in \mathcal{R}^{H\times W}$ and $\mathbf{M}_{TS}\in \mathcal{R}^{H\times W}$.

\textbf{Dual-Branch SalStructIQA Network.} In this architecture, there are two parallel branches of CNN equipped with a global average pooling layer(GAP) for feature learning, as shown in Fig. \ref{fig:architecture}(a) . One branch takes the combination of $\mathbf{I}$ and $\mathbf{M}_{LS}$ as input and produces a feature vector denoted as $\mathbf{f}_{LS}$. The other branch takes the combination of $\mathbf{I}$ and $\mathbf{M}_{TS}$ as input and produces a feature vector denoted as $\mathbf{f}_{TS}$. Then these two feature vectors are concatenated and fed into a full connection (FC) layer to predict the quality grade:
\begin{equation}
	\hat{y} =\mathrm{softmax}\left( \mathbf{w}^{T} [\mathbf{f}_{LS}, \mathbf{f}_{TS}] + \mathbf{b}\right)\;,
\end{equation}
where $\mathbf{w}$ and $\mathbf{b}$ are weights and bias to be learned in FC layer. Parameters in both parallel CNN backbones and FC layers are learned in an end-to-end way and cross-entropy loss is used  in training phase.

One thing worth noting is that arbitrary existing CNN backbones such as VGGNet, ResNet and DenseNet etc. can be used for feature learning in each branch in Fig. \ref{fig:architecture}(a) after revising the first dimension of convolution kernel in the first convolutional layer to four. In our experiments, performances of two famous backbones, i.e., ResNet-50 and DenseNet-121 are reported. 

\textbf{Single-Branch SalStructIQA Network.} The architecture is shown in Fig. \ref{fig:architecture}(b). It consists of one CNN branch together with a GAP for feature learning and a FC layer for classification. This architecture takes $[\mathbf{I}, \mathbf{f}_{LS}, \mathbf{f}_{TS}]$ as an input and predicts the quality grade. Same to dual-branch one, the network can be trained in an end-to-end way and cross-entropy loss is used in training phase.


\section{Experimental results}
\label{sec:exprst}
The proposed SalStructIQA method is first evaluated and compared with the state-of-the-art image quality assessment methods on the public dataset Eye-Quality \cite{fu2019evaluation}. Eye-Quality \cite{fu2019evaluation} collects 28,792 retinal images from Kaggle and each retinal image is re-annotated with three grades \lq Good\rq, \lq Usable\rq~and \lq Reject\rq. The image size ranges from $211 \times 320$ to $3456\times 5184$. Those images are divided into into two subsets: 12,543 for training and 16,249 for testing. Then extensive ablation study on Eye-Quality \cite{fu2019evaluation} is conducted. Following \cite{fu2019evaluation}, average accuracy ($Acc$), precision ($P$), recall ($R$) and F-score ($F_{score}=2\times P \times R /(P+R)$) are used as evaluation metrics.



\subsection{Data augmentation and experimental setup}
\textbf{Data preprocessing and augmentation.} With the retinal images, we follow \cite{fu2019evaluation} and first crop the field of view (FoV) via  Hough Circle Transform, then we pad short side with zeros such that the width and height of the cropped FoV regions are equal length. Finally, we rescale padded regions to $224\times 224$. The masks of large-sized salient structures $\mathbf{M}_{LS}$ and tiny-sized salient structures $\mathbf{M}_{TS}$ are detected from rescaled images. To increase the diversity of training data, we use vertical flipping, horizontal flipping and random rotation to augment the training data.

\textbf{Experimental setup.} Parameters in backbone network are initialized with the pre-trained models on ImageNet \cite{deng2009imagenet}.  Parameters associated with full connection layers are initialized with Gaussian distribution with zero mean and standard deviation 0.001. We adopt stochastic gradient descent to optimize the network with initial learning rate of $0.01$. We train the model 20 epochs and reduce the learning rate to 0.001 after 10 epochs. 

\begin{table*}[ht]\scriptsize
\centering
\caption{ Comparisons of the proposed methods and state-of-the-arts on Eye-Quality \cite{fu2019evaluation}}
\label{Tab:ComparisonSOTA}
\begin{tabular}{c|c|c|c|c|c|c}
\hline
Methods&   & $Acc$ & $P$ & $R$ & $F_{score}$ & $Param$\\
\hline
Natural-IQA & BRISQUE\cite{mittal2012no} & 0.7692& 0.7608 & 0.7095 & 0.7112 & -\\
 & NBIQA\cite{Fu2019A} & 0.7917& 0.7641 & 0.7509 & 0.7441  & -\\
& TS-CNN \cite{yan2018two} & 0.7926& 0.7976 & 0.7446 & 0.7481 & 1.44M\\
\hline
Retinal-IQA & HVS-based* \cite{wang2015human}& - & 0.7404 & 0.6945 & 0.6991 & -\\
 & MR-CNN* \cite{raj2020multivariate} & 0.8843 & \underline{0.8697} & \underline{0.8700} & \underline{0.8694} & 101.80M\\
  & DenseNet121-MCF* \cite{fu2019evaluation} & - & 0.8645 & 0.8497 & 0.8551 & 28.26M\\
   & DenseNet121-MCF \cite{fu2019evaluation} & 0.8722 & 0.8563 & 0.8482 & 0.8506& 28.26M\\ 
   & DenseNet121-RGB* \cite{fu2019evaluation} & - & 0.8194 & 0.8114 & 0.8152 &6.96M\\
   & DenseNet121-RGB \cite{fu2019evaluation} & 0.8568 & 0.8481 & 0.8239 & 0.8315&6.96M\\
 & Single-branch SalStructIQA & \underline{0.8847} & \underline{0.8715} & 0.8645 & 0.8662 &6.96M \\
 & Dual-branch SalStructIQA & \textbf{0.8897} & \textbf{0.8748} & \textbf{0.8721} & \textbf{0.8723} &13.92M\\
\hline
\end{tabular}
\end{table*}

\subsection{Comparisons with State-of-the-arts}
We compare our SalStructIQA with six image quality assessment methods: BRISQUE \cite{mittal2012no}, NBIQA \cite{Fu2019A} and TS-CNN \cite{yan2018two}, HVS-based method \cite{wang2015human}, DenseNet121-MCF \cite{fu2019evaluation} and Multivariate-Regression CNN (MR-CNN) \cite{raj2020multivariate}. The first three methods are designed for Natural-IQA, among which BRISQUE \cite{mittal2012no} and NBIQA \cite{Fu2019A} are hand-crafted feature based and TS-CNN \cite{yan2018two} is deep feature based. Their results are obtained by adapting the codes from authors to Retinal-IQA task with Eye-Quality dataset \cite{fu2019evaluation}. In detail, those three Natural-IQA methods aim to regress the quality scores of natural images while Retinal-IQA in this paper aims to classify the quality of retinal images into three classes. Thus to make those Natural-IQA methods adapt to Retinal-IQA, regression is replaced with classification. Particularly, in the implementations of BRISQUE \cite{mittal2012no} and NBIQA \cite{Fu2019A}, we directly predict retinal image quality label instead of regressing the quality score. They are implemented by Matlab 2014A on the platform of a PC with an i5 CPU and 16 GB RAM. In the implementation of TS-CNN \cite{yan2018two}, the regression layer with one-dimensional output is replaced with a classification layer with three-dimensional outputs and the loss is replaced by multi-class cross-entropy loss. The last three methods are specifically designed for Retinal-IQA, among which HVS-based method \cite{wang2015human} is hand-crafted feature based and  DenseNet121-MCF \cite{fu2019evaluation} and MR-CNN \cite{raj2020multivariate} are deep feature based. The results of HVS-based method \cite{wang2015human} and DenseNet121-MCF \cite{fu2019evaluation} are from \cite{fu2019evaluation}. For fair comparison, the $Acc$ results of HVS-based method \cite{wang2015human} and DenseNet121-MCF \cite{fu2019evaluation} are not given here because they do not report their corrected result. The results of MR-CNN \cite{raj2020multivariate} are directly taken from the original paper. We also reproduce the results of DenseNet121-MCF \cite{fu2019evaluation} and DenseNet121-RGB \cite{fu2019evaluation}. 

Table. \ref{Tab:ComparisonSOTA} reports the performances comparing with state-of-the-arts. Results from the original papers or reproduced by others are marked with * while those without any mark are reproduced by ours. Our reproduction of previous methods and implementation of our methods are conducted with three independent trails and the average performances are reported.  From Table. \ref{Tab:ComparisonSOTA}, we have following observations.
\begin{itemize}
	\item  Comparing with hand-crafted feature based methods, i.e. BRISQUE \cite{mittal2012no}, NBIQA \cite{Fu2019A} and HVS-based method \cite{wang2015human}, deep feature based methods achieve better performances on Retinal-IQA. The possible reason is that deep features are more powerful than hand-crafted features.
	
	
	\item For deep feature based methods,  Retinal-IQA methods  significantly outperforms Natural-IQA methods, i.e. TS-CNN \cite{yan2018two}.
	  
	\item Comparing with deep features based Retina-IQA method MR-CNN* \cite{raj2020multivariate}, our Single-branch SalStructIQA achieves 88.47\% in $Acc$ which slightly outperforms MR-CNN* \cite{raj2020multivariate} by 0.04\%. In terms of $F_{score}$, our Single-branch SalStructIQA achieves 86.62\%, which is slightly inferior to MR-CNN* \cite{raj2020multivariate} by 0.31\%. One thing worth to mention is that the parameters of our Single-branch SalStructIQA are approximately $1/15$ of MR-CNN* \cite{raj2020multivariate}. 
	
	\item Comparing with DenseNet121-MCF* \cite{fu2019evaluation}, our Single-branch SalStructIQA outperforms it by 1.11\% in $F_{score}$. Additionally, our Single-branch SalStructIQA is much lighter than DenseNet121-MCF* \cite{fu2019evaluation}, whose parameter scale is approximately $1/4$ of DenseNet121-MCF* \cite{fu2019evaluation}. Comparing with DenseNet121-RGB* \cite{fu2019evaluation}, ours Single-branch SalStructIQA significantly outperforms it with the same parameter scale.	
	
	\item Comparing with the Single-branch one, our Dual-branch SalStructSal further improves the performances consistently while the parameter scale is double. Comparing with both Natural-IQA methods and Retinal-IQA methods listed in Table. \ref{Tab:ComparisonSOTA}, our Dual-branch SalStructIQA achieves best performances. 
\end{itemize}

\begin{figure*}[h]
	\centering
	\includegraphics[width=0.95\linewidth]{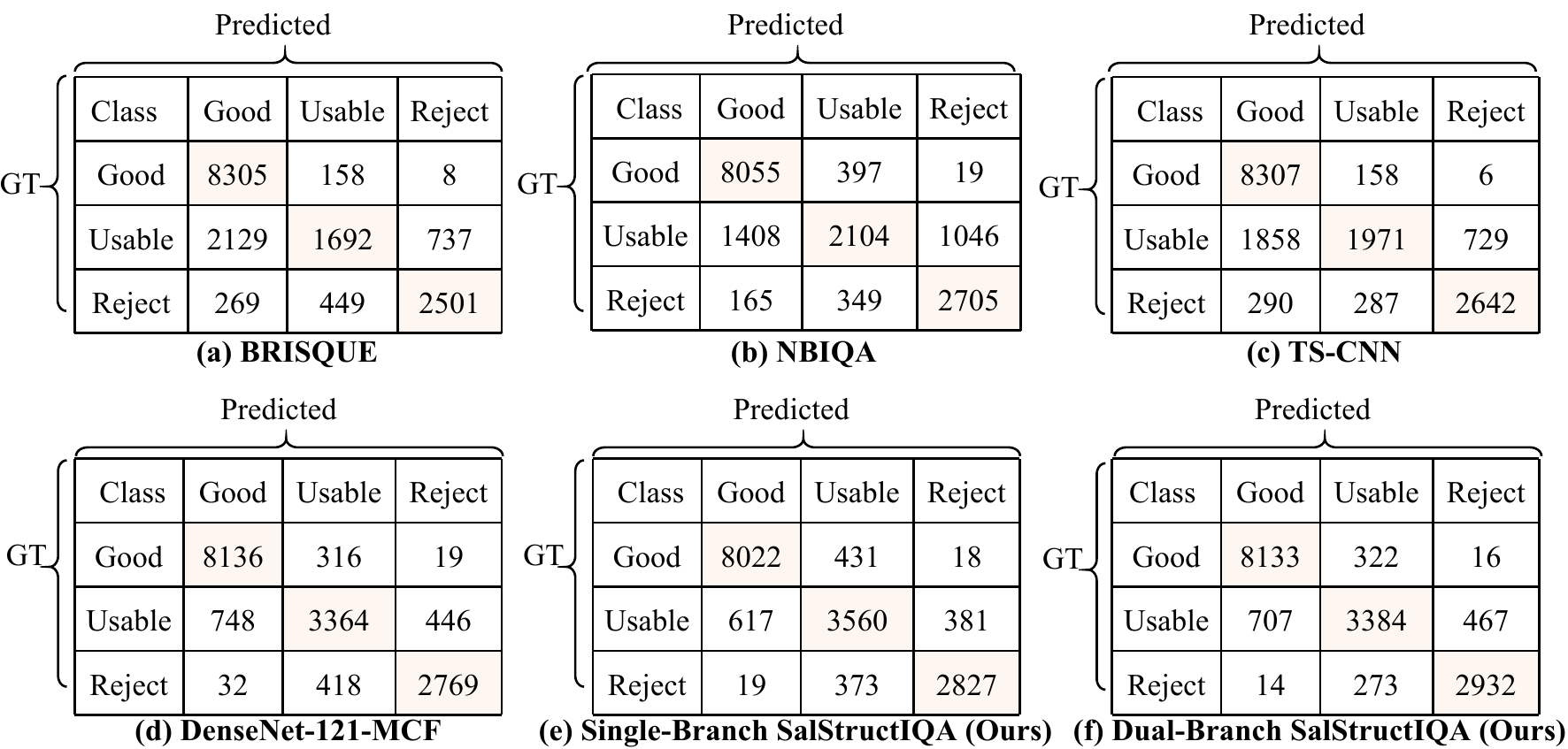}
	\caption{Confusion matrices of BRISQUE \cite{mittal2012no}, NBIQA \cite{Fu2019A}, TS-CNN\cite{yan2018two}, DenseNet121-MCF \cite{fu2019evaluation}, our Single-branch SalStructIQA and Dual-branch SalStructIQA.}
	\label{fig:ConfusionMatrixSOTA}
\end{figure*}
  
We also report the confusion matrices of methods BRISQUE \cite{mittal2012no}, NBIQA \cite{Fu2019A}, TS-CNN\cite{yan2018two}, DenseNet121-MCF \cite{fu2019evaluation}, our Single-branch SalStructIQA and Dual-branch SalStructIQA in Fig. \ref{fig:ConfusionMatrixSOTA}. For brevity, we report the confusion matrices of  those methods according to the trail achieving best $F_{score}$. From confusion matrices we have following observations:
\begin{itemize}
	\item Natural-IQA methods BRISQUE \cite{mittal2012no}, NBIQA \cite{Fu2019A} and TS-CNN\cite{yan2018two} work well in classifying retinal images with \lq Good\rq~quality. But they tend to wrongly classify retinal images graded as \lq Usable\rq~into \lq Good\rq~and \lq Reject\rq~and classify retinal images graded as \lq Reject\rq~into \lq Good\rq~and \lq Usable\rq.
	
	\item Comparing with Natural-IQA methods, DenseNet121-MCF \cite{fu2019evaluation} improves the classification  accuracy of retinal images graded as \lq Usable\rq.
	
	\item Comparing with Natural-IQA methods, both our Single-branch method and Dual-branch method achieve higher classification accuracies of retinal images graded as \lq Usable\rq~and \lq Reject\rq~significantly. Comparing with DenseNet121-MCF \cite{fu2019evaluation}, ours method performs better in classification on retinal images graded as \lq Reject\rq, which is more important than correctly identifying the retinal images graded as \lq Usable\rq~and \lq Good\rq~in clinical.
\end{itemize}

\label{fig:mscn_retinal}
\begin{table*}[ht]\scriptsize
	\centering
	\caption{The influences of backbone networks and our proposed salient structure priors. }
	\label{tab:AblationStudy}
	\begin{tabular}{c|c|cccc}
		\hline
	Model & Backbone& $Acc$ & $P$ & $R$ & $F_{score}$ \\ \hline\hline
Baseline & ResNet50   & 0.8809 & 0.8670 & 0.8616 & 0.8609 \\ \hline
+LS & ResNet50   & 0.8834 & 0.8626 & 0.8703 & 0.8642 \\ \hline
+TS & ResNet50   & 0.8833& 0.8663 & 0.8661 & 0.8646  \\ \hline
+LS+TS (Single-branch)&  ResNet50 & 0.8840& 0.8656 & 0.8680 & 0.8657 \\ \hline	
+LS+TS (Dual-branch) & ResNet50   & \textbf{0.8857}& \textbf{0.8694} & \textbf{0.8708} & \textbf{0.8695}  \\ \hline\hline
Baseline		& DenseNet121  & 0.8834 & 0.8711 & 0.8624 & 0.8641  \\	\hline
+		LS&  DenseNet121 & 0.8851 & 0.8706 & 0.8661 & 0.8672  \\\hline
+TS&DenseNet121 & 0.9229 & 0.8843 & 0.8631 & 0.8653 \\\hline
+LS+TS (Single-branch)&  DenseNet121 & 0.8847& 0.8715 & 0.8645 & 0.8662 \\ \hline		
+LS+TS (Dual-branch)&  DenseNet121 & \textbf{0.8897}& \textbf{0.8748} & \textbf{0.8721} & \textbf{0.8723}  \\
		\hline
	\end{tabular}
\end{table*}

\subsection{Ablation Study}
We further look into the influences of different backbone networks and our two proposed salient structures, i.e., large-size salient structures and tiny-size salient structures. To this end, ablation study is first conducted. Two prevalent backbone networks ResNet50 \cite{he2016deep} and DenseNet121 \cite{huang2017densely} are chosen. For each backbone network, we separately train four models: the baseline model which is trained with $\mathbf{I} \in \mathcal{R}^{H\times W \times 3}$, the model denoted as \lq+LS\rq~ which is trained with $[\mathbf{I}, \mathbf{M}_{LS}] \in \mathcal{R}^{H\times W \times 4}$, the model denoted as \lq+TS\rq~ which is trained with  $[\mathbf{I}, \mathbf{M}_{TS}] \in \mathcal{R}^{H\times W \times 4}$ and the model denoted as \lq+LS+TS (Single-branch)\rq~ which is trained with $[\mathbf{I}, \mathbf{M}_{LS}, \mathbf{M}_{TS}] \in \mathcal{R}^{H\times W \times 5}$. As mentioned before, $\mathbf{I}$, $\mathbf{M}_{LS}$ and $\mathbf{M}_{LS}$  are the RGB retinal image, the mask of large-size salient structures and the mask of tiny-size salient structures respectively. The performances are reported in Table \ref{tab:AblationStudy}. The performances of Dual-branch SalStructureIQA are also listed. Obviously, DenseNet121 \cite{huang2017densely} outperforms ResNet50 \cite{he2016deep} in Retinal-IQA. Moreover, from Table. \ref{tab:AblationStudy} we can see that (1) combining RGB retinal image with either large-size salient structures or tiny-size salient structures is able to improve the performances of Retinal-IQA, (2) our Dual-branch SalStructIQA achieves superior performances than the Single-branch one which indicates that separately learning features from $[\mathbf{I}, \mathbf{M}_{LS}]$  and $[\mathbf{I}, \mathbf{M}_{TS}]$ works better than learning features from $[\mathbf{I},\mathbf{M}_{LS}, \mathbf{M}_{TS}]$.

\begin{figure}[!ht]
	\centering
	\includegraphics[width=\linewidth]{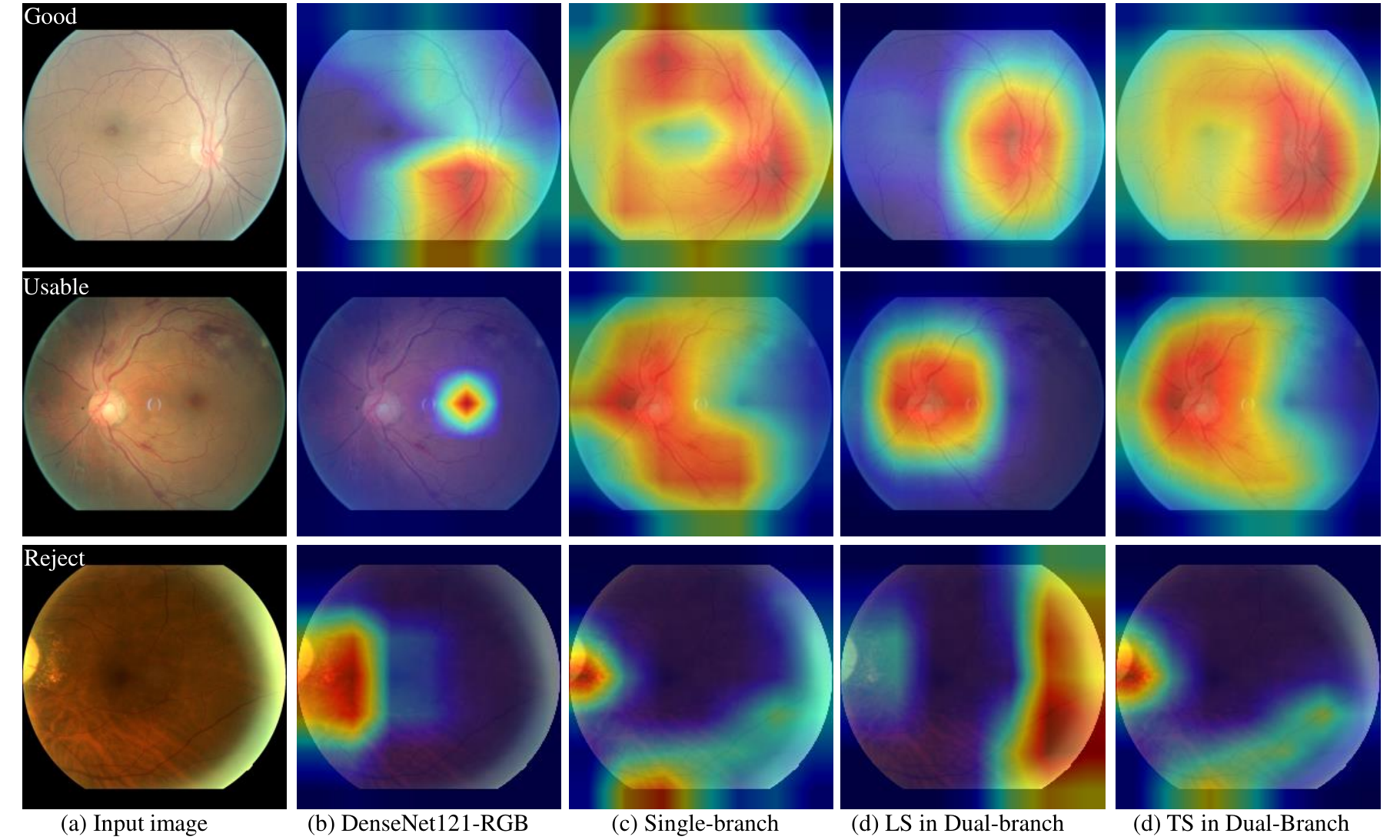}
	\caption{Examples of gradient-weighted class activation maps of different models generated according to \cite{selvaraju2017grad}. These three retinal images are mis-classified by DenseNet121-RGB \cite{fu2019evaluation} but correctly classified by our Single-branch SalStructIQA and Dual-branch SalStructIQA. Obviously, with salient structures, the models tend to learn discriminate information from salient regions for image quality assessment.}
	\label{fig:CAM}
\end{figure}

Additionally, we show three examples of gradient-weighted class activation maps of three models, i.e., DenseNet121-RGB \cite{fu2019evaluation}, our Single-branch SalStructIQA and Dual-branch SalStructIQA in Fig. \ref{fig:CAM}. Those gradient-weighted activation maps are generated according to \cite{selvaraju2017grad}. As our Dual-branch SalStructIQA contains two branches, i.e., the large-size salient structure branch and the tiny-size salient structure branch, thus for each input there are two class activation maps as listed in  the last two columns in Fig. \ref{fig:CAM}. With regard to these three examples graded as \lq Good\rq, \lq Usable\rq~and \lq Reject\rq~from top to bottom, DenseNet121-RGB \cite{fu2019evaluation} wrongly classifies them into \lq Usable\rq, \lq Good\rq~and \lq Usable\rq~while our two SalStructIQA models are able to correctly classify them. Gradient-weighted class activation maps of DenseNet121-RGB \cite{fu2019evaluation} (the second column in Fig. \ref{fig:CAM}) indicate that DenseNet121-RGB may fail to capture the most informative information for classification. On the contrary, our two models are able to capture the most informative information. As shown in the third column in Fig. \ref{fig:CAM}, for the classification of images graded as \lq Good\rq~and \lq Usable\rq, our Single-branch SalStructIQA are able to capture informative information from optic disc and vessels. For the classification of image graded as \lq Reject\rq, our Single-branch SalStructIQA are able to capture distinct information from the distortion regions, i.e. partial optic disc region and region with uneven illumination. As shown in the fourth and fifth columns in Fig. \ref{fig:CAM}, the large-size salient structure branch in our Dual-branch SalStructIQA is able to capture informative information from optic disc region for the classification of \lq Good\rq~and \lq Usable\rq~images and from the distorted bright region for the classification of \lq Reject\rq~image. The tiny-size salient structure branch is guided to focus on the vessels for Retinal-IQA. Comparing with DenseNet121-RGB \cite{fu2019evaluation} without considering the salient structures in retinal images, the ways that our proposed two models with salient structures for Retinal-IQA are much more consistent to that ophthalmologists perform quality assessment.


\section{Conclusion}
\label{sec:conclusion}

In this paper, we propose two salient structures priors, i.e., large-size salient structures and tiny-size salient structures for retinal image quality assessment. They are incorporated with deep CNN such that the CNN is guided to learn informative information from those salient structures for quality grading. Correspondingly, two CNN architectures termed Dual-branch SalStructIQA and Single-branch SalStructIQA are developed. The former uses the two priors to separately guide the learning of two CNN branches and is able to achieve superior performances than the state-of-the-arts on Eye-Quality \cite{fu2019evaluation}. The latter concatenates two priors into one branch to guide the single branch CNN which is much light-weight compared with state-of-the-arts but still able to achieve competitive performances on Eye-Quality \cite{fu2019evaluation}. We visualize the class activation maps of both architectures which shows that our proposed methods grade the quality of retinal images in a similar way to ophthalmologists.

\section{Acknowledgment}
This work is supported by the National Key Research and Development Program of China (No. 2018AAA0102102); and the National Natural Science Foundation of China (No. 62006249).





\bibliographystyle{elsarticle-num-names}
\bibliography{ref1}


\end{document}